%% file: template.tex
\documentclass{article}

\usepackage{arxiv}

\usepackage[utf8]{inputenc} 
\usepackage[T1]{fontenc}    
\usepackage{hyperref}       
\usepackage{url}            
\usepackage{booktabs}       
\usepackage{amsfonts}       
\usepackage{nicefrac}       
\usepackage{microtype}      
\usepackage{lipsum}		    
\usepackage{graphicx}
\usepackage{pifont}
\usepackage[sort,numbers]{natbib}
\usepackage{doi}
\usepackage{eso-pic}

\usepackage{subcaption}
\usepackage{numprint}
\usepackage[inline]{enumitem}
\usepackage[binary-units=true]{siunitx}
\usepackage[]{booktabs}
\usepackage[version=3]{mhchem}

\usepackage[figuresright]{rotating}
\usepackage{acro}
\usepackage{xcolor}

\hypersetup{
pdftitle={Out of Distribution Detection for Efficient Continual Learning in Quality Prediction for Arc Welding},
pdfauthor={Yannik Hahn, Jan Voets, Antonin Königsfeld, Hasan Tercan, Tobias Meisen},
pdfkeywords={Out-of-Distribution Detection, Continual Learning, Predictive Quality, Arc Welding, VQ-VAE Transformer, Time Series}
}

\AddToShipoutPictureBG*{%
  \AtPageUpperLeft{%
    \put(\LenToUnit{2cm},\LenToUnit{-1cm}){%
      \parbox{\textwidth}{%
        \footnotesize
        \textcopyright\ 2025 ACM. This is the author's version of the work. It is posted here for your personal use. Not for redistribution. The definitive Version of Record was published in \textit{Proceedings of the 34th ACM International Conference on Information and Knowledge Management (CIKM '25)}, \url{https://doi.org/10.1145/3746252.3761535}.
      }%
    }%
  }%
}

\title{Out of Distribution Detection for Efficient Continual Learning in Quality Prediction for Arc Welding}
\renewcommand{\shorttitle}{OOD Detection for Efficient Continual Learning in Quality Prediction for Arc Welding}

\author{
\href{https://orcid.org/0000-0003-4046-2990}{\includegraphics[scale=0.06]{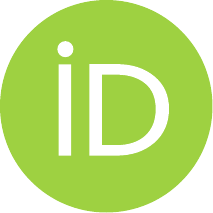}\hspace{1mm}Yannik Hahn}\thanks{Corresponding author. {\it E-mail address:} yhahn@uni-wuppertal.de}, 
\href{https://orcid.org/0009-0004-4542-1017}{\includegraphics[scale=0.06]{orcid.pdf}\hspace{1mm}Jan Voets}, \href{https://orcid.org/0009-0009-2757-9570}{\includegraphics[scale=0.06]{orcid.pdf}\hspace{1mm}Antonin Königsfeld}, 
\href{https://orcid.org/0000-0003-0080-6285}{\includegraphics[scale=0.06]{orcid.pdf}\hspace{1mm}Hasan Tercan}, 
\href{https://orcid.org/0000-0002-1969-559X}{\includegraphics[scale=0.06]{orcid.pdf}\hspace{1mm}Tobias Meisen} \\
\textit{{yhahn, voets, koenigsfeld, tercan, meisen}@uni-wuppertal.de} \\
Institute for Technologies and Management of Digital Transformation (TMDT)\\ 
University of Wuppertal \\
Rainer-Gruenter-Straße 21, 42119 Wuppertal, Germany \\ 
}

\date{Preprint, August 2025}
\renewcommand{\headeright}{A Preprint}
\renewcommand{\undertitle}{Accepted at CIKM 2025 (Applied Research Papers)}

\input{00_acronym}
\begin{document}
\maketitle

\input{01_abstract}

\input{02_introduction}
\input{03_related_work}
\input{04_use_case}
\input{05_methodology}
\input{06_experiments}

\input{07_conclusion_outlook}
\input{08_gen_ai_statement}

\bibliographystyle{plainnat}
\bibliography{ref}  

\end{document}

%% file: 00_acronym.tex
\DeclareAcronym{AL}{ short = AL, long = arc length }
\DeclareAcronym{VQ-VAE}{ short = VQ-VAE, long = vector quantised variational autoencoder }
\DeclareAcronym{VQ}{ short = VQ, long = vector quantisation }
\DeclareAcronym{VAE}{ short = VAE, long = variational autoencoder }
\DeclareAcronym{CNN}{ short = CNN, long = convolutional neural network }
\DeclareAcronym{DIN}{ short = DIN, long = German Institute for Standardisation Registered Association }
\DeclareAcronym{DNP}{ short = DNP, long = distance from nozzle to plate }
\DeclareAcronym{EWC}{ short = EWC, long = elastic weight consolidation }
\DeclareAcronym{GFR}{ short = GFR, long = gas flow rate }
\DeclareAcronym{GMAW}{ short = GMAW, long = gas metal arc welding }
\DeclareAcronym{HIVE-COTE}{ short = HIVE-COTE, long =  }
\DeclareAcronym{LSTM}{ short = LSTM, long = long short-term memory }
\DeclareAcronym{MAG}{ short = MAG, long = metal active gas }
\DeclareAcronym{MAS}{ short = MAS, long = memory-aware synapses }
\DeclareAcronym{MIG}{ short = MIG, long = metal inert gas }
\DeclareAcronym{RNN}{ short = RNN, long = recurent neural network }
\DeclareAcronym{VPRS}{ short = VPRS, long = variable precision rough set }
\DeclareAcronym{WFR}{ short = WFR, long = wire feed rate }
\DeclareAcronym{WS}{ short = WS, long = welding speed }
\DeclareAcronym{ml}{ short = ML, long = machine learning }
\DeclareAcronym{dl}{ short = DL, long = deep learning }
\DeclareAcronym{MLP}{ short = MLP, long = multilayer perceptron }
\DeclareAcronym{GRU}{ short = GRU, long = gated recurrent unit }
\DeclareAcronym{CV}{ short = CV, long = computer vision }
\DeclareAcronym{NLP}{ short = NLP, long = natural language processing }
\DeclareAcronym{CMT}{ short = CMT, long = cold metal transfer }
\DeclareAcronym{VQ-VAE-Tr}{ short = VQ-VAE Tr, long = VQ-VAE Transformer }

%% file: 01_abstract.tex
\begin{abstract}
Modern manufacturing relies heavily on fusion welding processes, including \ac*{GMAW}. Despite significant advances in machine learning-based quality prediction, current models exhibit critical limitations when confronted with the inherent distribution shifts that occur in dynamic manufacturing environments. In this work, we extend the VQ-VAE Transformer architecture—previously demonstrating state-of-the-art performance in weld quality prediction—by leveraging its autoregressive loss as a reliable out-of-distribution (OOD) detection mechanism. Our approach exhibits superior performance compared to conventional reconstruction methods, embedding error-based techniques, and other established baselines. By integrating OOD detection with continual learning strategies, we optimize model adaptation, triggering updates only when necessary and thereby minimizing costly labeling requirements. We introduce a novel quantitative metric that simultaneously evaluates OOD detection capability while interpreting in-distribution performance. Experimental validation in real-world welding scenarios demonstrates that our framework effectively maintains robust quality prediction capabilities across significant distribution shifts, addressing critical challenges in dynamic manufacturing environments where process parameters frequently change. 
This research makes a substantial contribution to applied artificial intelligence by providing an explainable and at the same time adaptive solution for quality assurance in dynamic manufacturing processes - a crucial step towards robust, practical AI systems in the industrial environment.
\end{abstract}

%% file: 02_introduction.tex
\section{Introduction}
Arc welding processes, particularly \ac{GMAW}, form the backbone of modern manufacturing across automotive, marine, and aerospace industries~\cite{Bestard.2018}. Despite its widespread adoption, \ac{GMAW} presents significant challenges stemming from the complex interplay between process parameters—current, voltage, welding speed—and resultant weld quality. Traditional quality assessment approaches rely heavily on operator expertise and post-process inspection, limiting real-time quality prediction capabilities. These challenges can be effectively addressed through machine learning and deep learning methodologies, which leverage sensor and process data to develop predictive models capable of real-time quality assessment.

Manufacturing environments are inherently dynamic, with frequent changes in setup, materials, and process parameters creating distribution shifts in sensor data. These shifts present significant challenges for machine learning models trained on static distributions. While continual learning offers a solution for adapting to evolving data without catastrophic forgetting, determining when to initiate model updates remains challenging. Frequent retraining is impractical due to the high cost of obtaining labeled data in manufacturing contexts, yet failing to detect distribution shifts leads to degraded model performance. Thus, maintaining robust quality prediction capabilities across evolving manufacturing environments represents the most persistent challenge in industrial machine learning deployments.

This work addresses the critical challenge of detecting out-of-distribution (OOD) data relative to a model's learned distribution, particularly during transitions between fundamentally different welding types (e.g., overlap to T-joints). Our work builds upon an existing state-of-the-art \ac{VQ-VAE-Tr} model for arc welding quality prediction~\cite{hahnCIKM}. While this model has advantages in the manufacturing domain through a combination of unsupervised and supervised learning, the architecture offers a high potential for OOD detection. We present an approach that leverages autoregressive loss as an effective OOD indicator, outperforming traditional methods (e.g. Maximum Softmax Probability (MSP)~\cite{hendrycks2018baselinedetectingmisclassifiedoutofdistribution} or ODIN~\cite{928a56b7d6f1473e930f282a0c4b534e}) as well as reconstruction- and quantization error-based detection approaches. This OOD detection mechanism serves as a trigger for initiating continual learning only when necessary, balancing adaptation needs with computational efficiency, reducing the costs of labeling data and using training resources.

Our approach is designed for real-world manufacturing scenarios where distribution shifts are inevitable but potentially unpredictable for a worker. These undetected shifts cause silent model degradation, where prediction accuracy deteriorates without warning, potentially resulting in undetected defects and costly production recalls. By integrating OOD detection with continual learning, we demonstrate improved model performance across significant distribution shifts while minimizing labeling requirements. This system establishes a framework for maintaining robust quality prediction capabilities across evolving manufacturing environments, thereby addressing one of the most persistent challenges in industrial machine learning deployments.

The main contributions of this paper are:
\begin{enumerate}
\item We demonstrate that autoregressive loss of a \ac{VQ-VAE-Tr} architecture serves as a superior OOD indicator compared to reconstruction or quantization errors and other baselines, providing a reliable trigger mechanism for model adaptation without requiring additional computational overhead.
\item We introduce a metric that measures the model’s ability to differentiate between in-distribution (ID) and out-of-distribution (OOD) data, and integrates the ID quality prediction performance.
\item We demonstrate in a real-world scenario that our method can be effectively combined with continual learning, thereby maintaining predictive performance across distribution shifts.
\end{enumerate}

%% file: 03_related_work.tex
\section{Related Work}
\subsection{Quality Prediction in Welding}
Welding quality prediction has evolved from traditional mathematical models requiring manual feature engineering to modern deep learning models that automatically learn representations from sensor data~\cite{10.1145/3340531.3412737, hahnCIKM}. Recent advances in machine learning, including neural networks of varying complexity and recurrent architectures, have improved predictive accuracy for welding performance~\cite{Zhang.2021, Kesse.2020, zhouPredictingQualityAutomated2020, 10.1145/3511808.3557512}. Among these, the \ac{VQ-VAE-Tr} architecture introduced by~\cite{hahnCIKM} achieved state-of-the-art performance in time series welding quality prediction by combining unsupervised and supervised learning approaches.

However, even these advanced approaches share a critical limitation: they assume static data distributions and lack the integration of mechanisms to detect or adapt to the distribution shifts inherent in dynamic manufacturing environments. As a result, these methods fail when welding configurations change—such as transitions from overlap to T-joints.

Our work addresses this gap by extending the \ac{VQ-VAE-Tr} architecture, leveraging its inherent but previously untapped potential for distribution shift detection through its multi-faceted representational capabilities. We develop robust OOD detection mechanisms that enable the system to identify manufacturing process variations and trigger appropriate model adaptation when necessary.

\subsection{OOD-Detection Methods}
Out-of-distribution (OOD) detection aims to identify data that deviates from a model's training distribution—a critical capability for maintaining reliability in dynamic environments. While OOD generalization focuses on maintaining performance across distribution shifts \cite{MustafaAbdool.2023, Lu.2022, Yang.2022}, OOD detection explicitly identifies when such shifts occur, enabling appropriate system responses. Existing OOD detection methods fall into three main categories, each with distinct advantages and limitations for manufacturing applications:

\subsubsection{Post-hoc methods} These approaches apply detection mechanisms to pre-trained models without modifying the training process. Classical approaches include MSP \cite{hendrycks2018baselinedetectingmisclassifiedoutofdistribution}, Energy Score~\cite{928a56b7d6f1473e930f282a0c4b534e}, and ODIN~\cite{liang2020enhancingreliabilityoutofdistributionimage}, which leverage output confidence calibration. Recent advances incorporate bias correction and entropy-based measures \cite{he2022outofdistributiondetectionunsupervisedcontinual}. While computationally efficient, these methods often struggle with temporal distribution shifts common in time series manufacturing data.

\subsubsection{Preprocessing methods} This category focuses on modifying input data before training to improve OOD sensitivity. These range from normalization techniques to sophisticated approaches like AdaRNN's temporal distribution characterization, which optimizes data splitting to maximize distributional dissimilarity during training \cite{Du.2021}. However, preprocessing methods require prior knowledge of potential distribution shifts—an unrealistic assumption in evolving manufacturing processes—making them unsuitable for our welding application where novel joint configurations and process parameters emerge unpredictably.

\subsubsection{Parametrized approaches} This class of methods integrates OOD detection directly into the model architecture or training process. Such methods either learn explicit OOD representations to generate detection scores \cite{Feng.2021, Kaur.2022, Fan.2022} or modify training objectives to maximize discrimination between in-distribution and OOD samples \cite{Aguilar.2023}. 

Within this category, generative model-based approaches have shown promise for OOD detection. Variational Autoencoders (VAEs) have been employed for OOD detection in several recent works \cite{NEURIPS2022_3066f60a, NEURIPS2020_eddea82a, osada2023out, nanaumi2024low}. However, traditional VAE approaches rely on continuous latent spaces and reconstruction-based anomaly scores, which can struggle to capture subtle temporal pattern deviations in time series data. The discrete tokenization in VQ-VAE architectures fundamentally changes this dynamic—by converting continuous sensor signals into discrete codes, we enable the transformer to model sequence probabilities through autoregressive prediction rather than reconstruction error, providing a more sensitive mechanism for detecting temporal distribution shifts characteristic of manufacturing processes.

Our work builds upon the generative modeling paradigm but addresses these limitations by leveraging the autoregressive loss of the transformer component within our \ac{VQ-VAE-Tr} architecture. Similar to approaches in medical imaging \cite{GRAHAM2023102967}, we exploit the inherent temporal modeling capabilities as a natural OOD indicator. This eliminates the need for separate detection modules while providing superior sensitivity to the temporal dependencies characteristic of welding process data, effectively bridging the gap between unsupervised OOD detection and supervised quality prediction tasks.

\subsection{OOD-Detection Evaluation}
Rigorous evaluation of out-of-distribution detection methods requires appropriate performance metrics that quantify a model's capability to differentiate between in-distribution and OOD samples across various operational contexts. The Area Under the Receiver Operating Characteristic curve (AUROC) is a widely used metric in OOD detection evaluation \cite{franc2023reject, tajwar2021no, fort2021exploring}. AUROC measures a detector's ability to discriminate between ID and OOD samples across various decision thresholds, providing a threshold-independent performance measure. However, the selected threshold is critical in real-world applications \cite{humblot2023beyond}. \citet{humblot2023beyond} introduces the AUTC (Area Under the Threshold Curve) metric for evaluating OOD detection performance, which focuses on score separability rather than requiring a specific threshold. The metric deliberately focuses on the binary ID/OOD distinction, independent of whether the model correctly classifies ID samples into their respective classes. Another metric is the False Positive Rate at a fixed True Positive Rate (e.g. FPR@95\%TPR in \cite{vapsi2025hypercone}). This practical metric measures how many OOD samples are incorrectly classified as ID when the model correctly identifies 95\% of ID samples relying on one arbitrary threshold. 

\subsection{OOD-Detection Methods for Timeseries}
While OOD detection has been extensively studied in computer vision \cite{he2022outofdistributiondetectionunsupervisedcontinual, Aguilar.2023, Kim.2022, Doorenbos.2024, Feng.2021}, time series applications remain significantly underexplored \cite{ramakrishna2021efficientoutofdistributiondetectionusing, 9797620, Kaur.2022}. This gap is particularly critical for manufacturing environments where temporal patterns encode essential process dynamics.
Most existing time series OOD methods operate at the point level, treating each timestamp independently without considering temporal dependencies \cite{Kaur.2022}. These approaches—analogous to frame-by-frame detection in video—fail to capture the sequential nature of manufacturing processes where abnormalities often manifest as temporal pattern deviations rather than isolated outliers. For welding quality prediction, where current and voltage patterns evolve over complete welding cycles, such point-based detection is fundamentally insufficient.
CODiT \cite{Kaur.2022} represents one of the first temporal-aware OOD detection methods, utilizing non-conformity scores from an inductive conformal anomaly detection (ICAD) framework. The approach employs multiple variational autoencoders to learn temporal-equivariant features, generating reconstruction-based non-conformity scores that capture temporal dependencies. While this marks progress toward temporal OOD detection, reconstruction-based methods struggle with subtle distributional shifts common in manufacturing. This is a limitation that our autoregressive approach addresses by modeling the probabilities of temporal sequences directly.

\subsection{OOD and Continual Learning}
While OOD detection in continual learning has been extensively studied in computer vision through task- and class-incremental paradigms \cite{Aguilar.2023, vandeven2019scenarioscontinuallearning}, time series applications also remain underexplored. This gap is particularly critical for manufacturing environments, where process parameters evolve continuously rather than through discrete task boundaries, fundamentally violating machine learning's stationary distribution assumption \cite{10.1145/3580305.3599880}. 

Manufacturing environments, particularly in welding processes, exhibit continuous evolution of operational parameters and system configurations, necessitating model adaptation strategies that mitigate catastrophic forgetting. While recent temporal adaptation methodologies, such as FSNet \cite{Pham.2022}, demonstrate promising capabilities in handling dynamic distributions, these approaches exhibit a critical limitation: the absence of explicit out-of-distribution detection mechanisms. This fundamental deficiency prevents the systematic determination of optimal retraining schedules, hindering the efficient allocation of resources in industrial machine learning deployments. The challenge lies in establishing quantitative criteria for identifying when model adaptation becomes imperative, while balancing computational costs against performance maintenance requirements.

%% file: 04_use_case.tex
\section{Application Context: Adaptive Quality Monitoring in Gas Metal Arc Welding}
\subsection{Gas Metal Arc Welding}
Arc welding is a specific type of fusion welding used to join metallic objects~\cite{Weman.2003}. In contrast to other welding techniques like gas or laser welding, arc welding utilizes electrical energy to create the heat necessary to melt welding wire and workpiece. Our use case lies in the field of \ac{GMAW}, a welding technique that uses a continuously fed welding wire as the electrode and shielding gas to protect the material from atmospheric influences. 

The procedurally and physically most important part of the welding setup is situated between the tip of the welding wire and the workpiece. When high voltage is applied between them, an electric arc forms as a result of the ionization of the injected protective gas. When the current density in the welding wire is sufficiently high, it begins to heat up and liquify. This consequently creates metal droplets that fall on the surface of the workpiece, creating a melt pool. With the progression of the welding, the melt pool dissipates its heat into the surrounding material, solidifies, and creates a weld seam that fuses the previously separated metal sheets \cite{hahn2023towards}.

\begin{figure}
    \centering
    \begin{subfigure}{0.41\linewidth}
        \includegraphics[width=\linewidth]{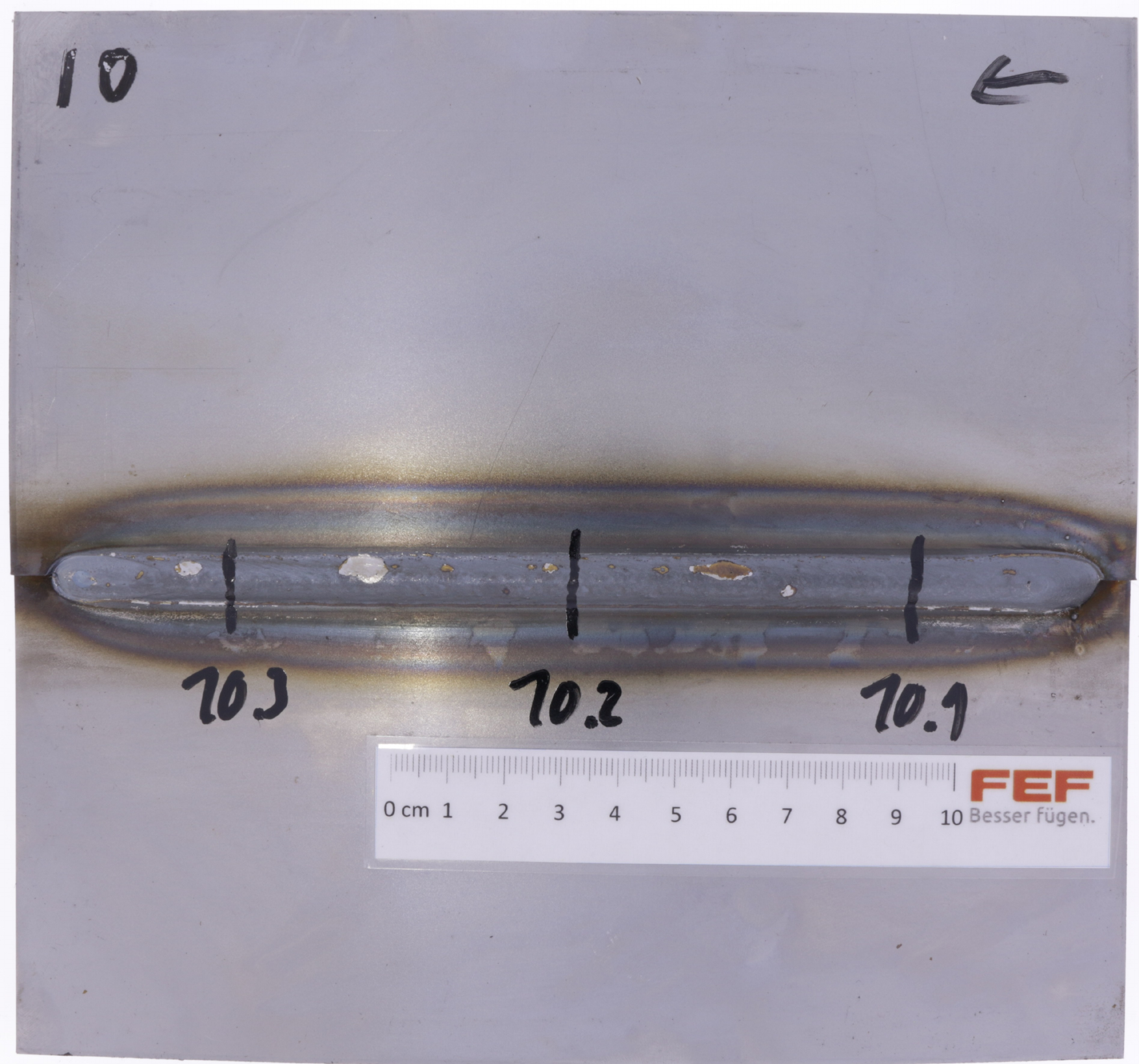}
        \caption{Overlap joint}
        \label{fig:sample_exp1}
    \end{subfigure}
    \hfill
    \begin{subfigure}{0.49\linewidth}
        \includegraphics[width=\linewidth]{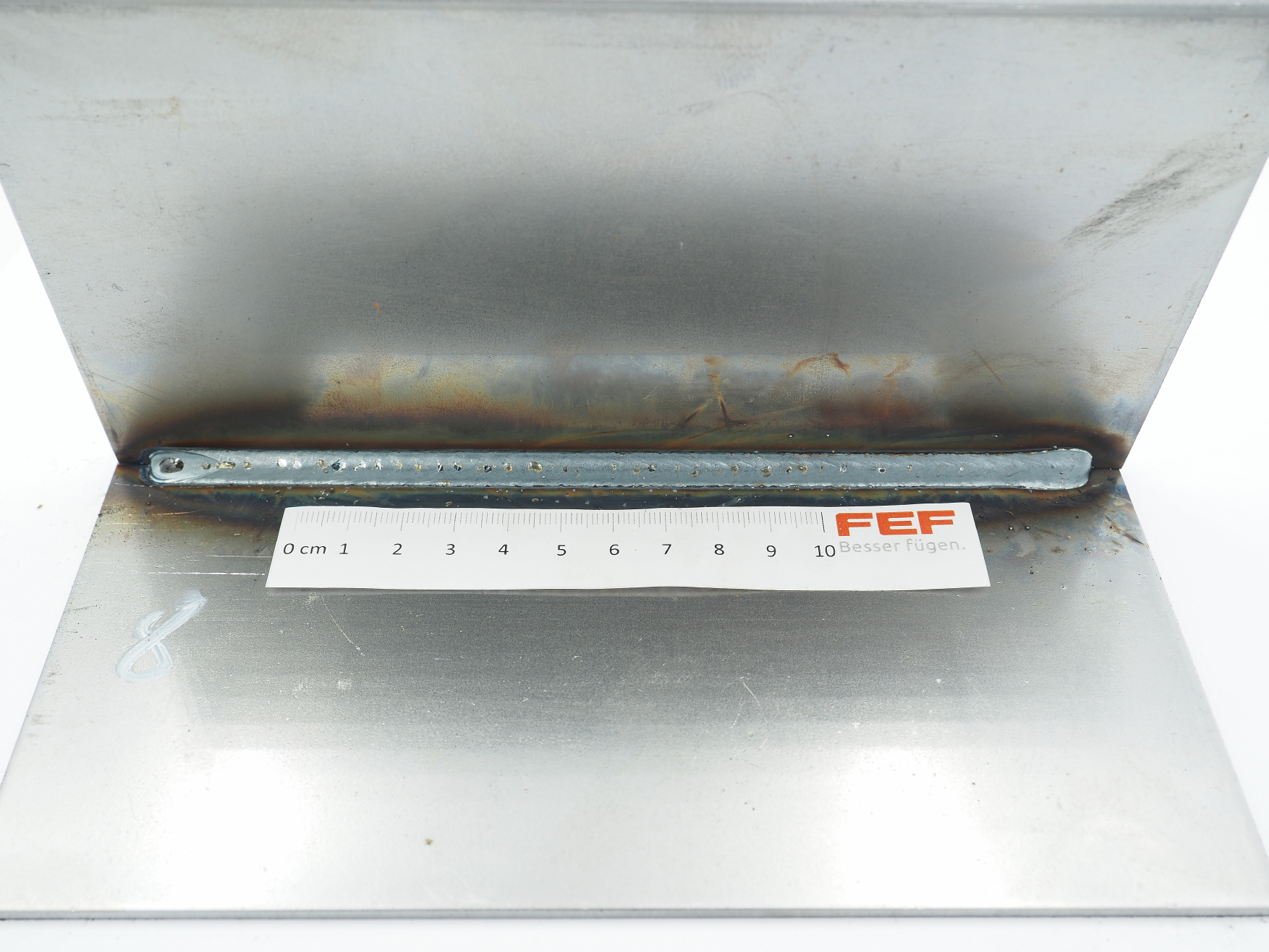}
        \caption{T-joint}
        \label{fig:sample_exp3}
    \end{subfigure}
    \caption{Sample images of the weld with different weld types.}
    \label{fig:sample_experiments}
\end{figure}

Critical insights into the welding process are provided by time series data, consisting of synchronized current and voltage measurements at high sampling rates (e.g., 100 kHz). Intricate local patterns and global interdependencies across multiple welding cycles (two welding cycle samples are shown in Figure~\ref{fig:sample_experiments}) are captured by these signals.

\subsection{Dataset}
\begin{figure}[thb!]
    \centering
    \begin{subfigure}{0.49\linewidth}
        \includegraphics[width=\linewidth]{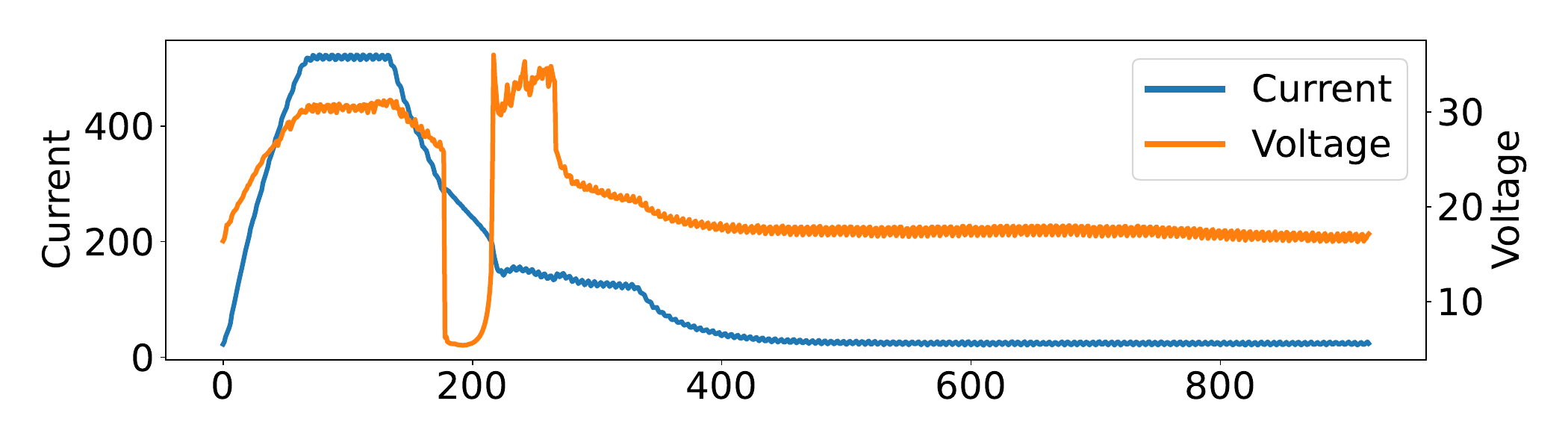}
    \end{subfigure}
    \hfill
    \begin{subfigure}{0.49\linewidth}
        \includegraphics[width=\linewidth]{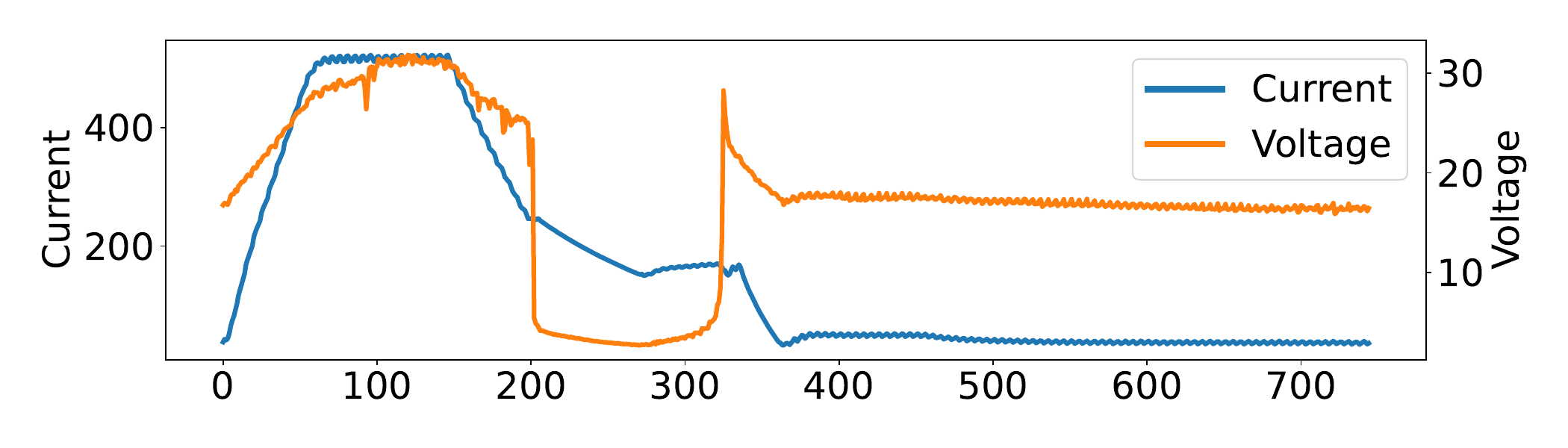}
    \end{subfigure}
    \caption{Representative current and voltage patterns from single welding cycles for overlap (left) and T-joint (right) configurations.}
    \label{fig:sample_plot_experimts}
\end{figure}

For our experimental validation, we utilized welding data acquired through multiple welding runs with systematically varied process control parameters. This dataset~\cite{hahn_2025_15497262} extends previous work by \citet{hahnCIKM} but is specifically structured to evaluate OOD detection capabilities across different joint configurations. The time-series data consists of current and voltage measurements segmented into welding cycles, as depicted in Figure\ref{fig:sample_plot_experimts}, creating a temporal classification problem.

Quality assessment in welding presents a unique challenge due to the destructive nature of evaluation methods. Following industry standards, quality labels were determined through metallographic cross-section examination at predefined intervals along the weld seam (Figure~\ref{fig:sample_experiments}). The labels follow a binary classification schema: substandard (0) and satisfactory (1), based on established welding quality standards.

\begin{table}[htb!]
    \centering
    \caption{Dataset partition designed to evaluate OOD detection capabilities across different joint configurations.}
    \begin{tabular}{r|c|c|c}
         \textbf{Dataset} &  \textbf{\#Samples} & \textbf{Satisfactory} & \textbf{Substandard}\\
         \midrule
         Train (Overlap Joint) &  \num{48758} &  \num{18547} & \num{30211} \\
         Val (Overlap Joint) &  \num{4676} &  \num{2488} & \num{2188 }\\
         Test (T-Joint) & \num{104531} & \num{58047} & \num{46484}\\
    \end{tabular}
    \label{tab:dist_class_data}
\end{table}

We conducted welding runs for two joint types — T--joints and overlap joints — which differ in their thermal dynamics and electrical characteristics. T-joints exhibit different heat dissipation patterns due to their perpendicular geometry, manifesting as subtle yet meaningful variations in current amplitude, voltage stability, and cycle periodicity.

The curated dataset encompasses \num{157965} labeled cycles. We allocated all overlap joint data to training and validation sets (maintaining temporal ordering), while reserving all T-joint data exclusively for testing (see Table~\ref{tab:dist_class_data}. This creates a distributional shift between training and testing conditions, enabling rigorous evaluation of our approach's capacity to detect OOD samples and trigger appropriate model adaptation.

%% file: 05_methodology.tex
\section{Methodology}

\subsection{Selective Adaptation via OOD-Triggered Continual Learning}
Common continual learning approaches in manufacturing have adopted a paradigm: updating models whenever any process parameter changes or a new task is defined, regardless of whether such changes impact the learned data distribution \cite{maschler2021regularization, tercan2022continual}. This indiscriminate adaptation strategy results in unnecessary computational overhead and, more critically, leads to substantial costs through redundant quality labeling procedures.

\begin{figure}
    \centering
    \includegraphics[width=1\linewidth]{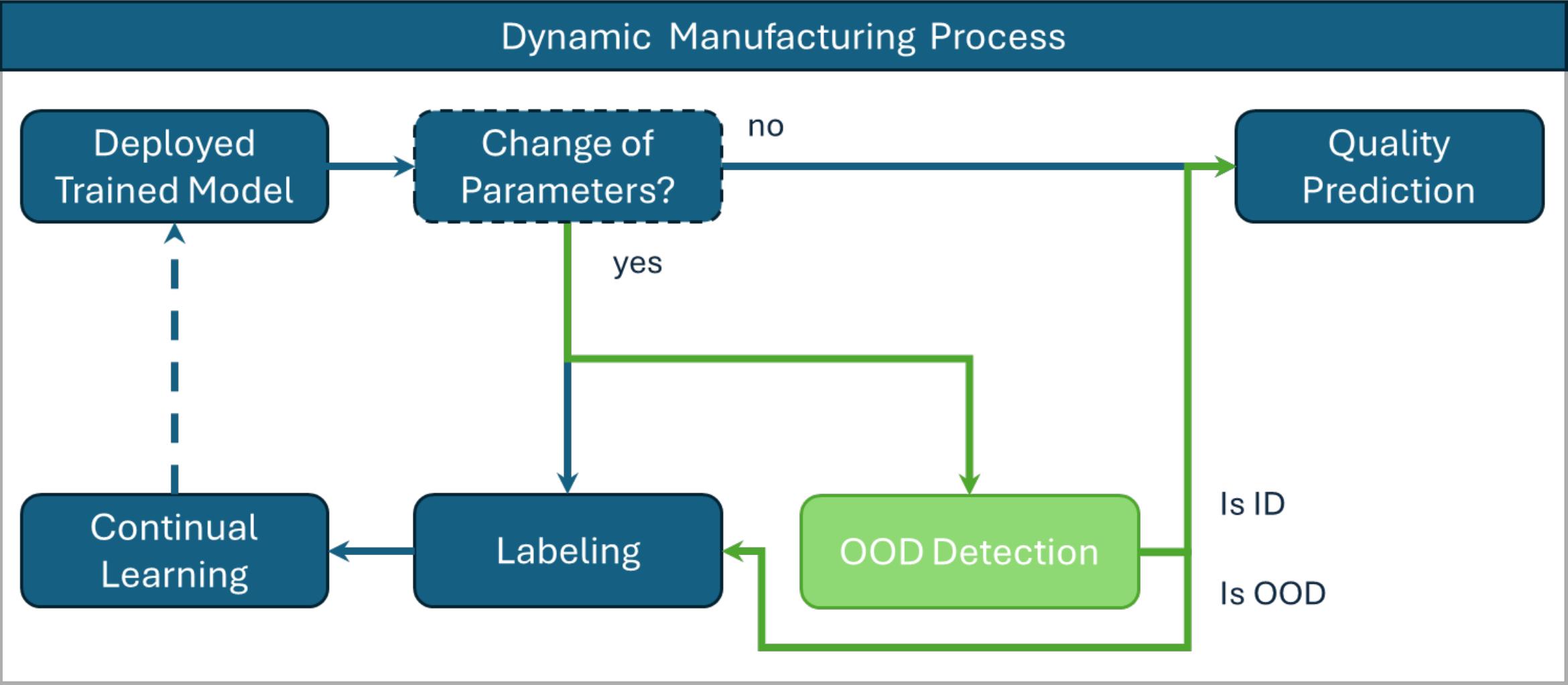}
    \caption{Selective adaptation framework using OOD detection (green pathway) as a trigger for continual learning, optimizing model updates only when process parameters change significantly, with the blue pathway showing the framework without OOD.}
    \label{fig:ood_cl_manufacturing}
\end{figure}

Our approach introduces an intelligent adaptation framework that employs OOD detection as a discriminative filter for continual learning activation. As illustrated in Figure~\ref{fig:ood_cl_manufacturing}, the framework continuously monitors incoming data streams and triggers model adaptation only when the OOD detector identifies samples that significantly deviate from the learned distribution. 

\subsection{OOD Score}
To evaluate model performance, we assess both ID prediction accuracy and OOD detection capability. For ID data, we use standard classification metrics (accuracy and F1-score) to ensure reliable quality predictions during deployment.

For OOD detection, we measure performance on an OOD test set. The ideal model performs poorly on OOD data—exhibiting low accuracy and F1-score—while maintaining high ID performance. This sharp contrast (maximized ID-OOD performance gap) directly indicates robust OOD detection capability.

Since effective models require both strong ID prediction and OOD detection, we introduce a unified OOD score that quantifies the model's ability to differentiate between ID and OOD data while incorporating ID prediction performance:

\begin{equation}
OOD-Score_{\rho} = \frac{\beta * ID_{\rho} + ID_{\rho} - OOD_{\rho}}{\beta * ID_{\rho} + ID_{\rho}}, \text{with } \rho \in \{acc, f1\}
\end{equation}
where \(\rho \in \{acc, f1\}\) denotes the evaluation metric (accuracy or F1-score) and \(\beta\) controls the weighting between ID performance and the ID-OOD gap. We set \(\beta=0.5\) to equally balance both objectives.

A high OOD score indicates strong ID performance coupled with effective OOD recognition (poor OOD performance). Conversely, a low score results from either: (i) poor ID prediction performance, or (ii) insufficient discrimination between ID and OOD data, where the model incorrectly maintains high performance on OOD samples.

\subsection{ROC-based Threshold Optimization}
Distinguishing between ID and OOD data requires establishing an appropriate decision boundary through systematic threshold determination. We implement a data-driven approach leveraging Receiver Operating Characteristic (ROC) curve analysis on validation data to derive optimal thresholds that are subsequently applied to test data.

For threshold determination, we analyze the distribution of OOD detection scores (e.g., autoregressive loss, MSP) across the validation data. This distribution typically exhibits a bimodal pattern: lower scores for correctly classified samples and higher scores for misclassified ones. The threshold acts as a "confidence boundary" within this distribution—samples with scores below it are considered reliable (ID), while those above are flagged as potentially unreliable (OOD). We identify this optimal boundary by analyzing where the score distribution best separates the model's correct predictions from its errors on validation data.

The threshold optimization follows a structured methodology centered on Youden's J statistic (TPR - FPR), which maximizes the vertical distance between the ROC curve and the diagonal line of no discrimination \cite{youden1950index}. This criterion provides an optimal trade-off between sensitivity and specificity by simultaneously maximizing the true positive rate while minimizing the false positive rate.

Our implementation proceeds through four discrete stages. First, we partition the validation set into correctly and incorrectly classified instances by comparing predicted labels against ground truth annotations. Second, we compute appropriate OOD indicators for each instance (e.g., MSP). For discriminative models, we extract confidence scores derived from maximum softmax probabilities, whereas for generative architectures like our \ac{VQ-VAE-Tr}, we utilize error metrics—specifically autoregressive loss.

Third, we construct the ROC curve by calculating the true positive and false positive rates across a spectrum of potential threshold values. Each point on the curve represents a specific threshold's performance in discriminating between correctly classified (presumed in-distribution) and incorrectly classified (potentially out-of-distribution) samples.

Finally, we identify the optimal threshold by locating the point on the ROC curve that maximizes the expression:

\begin{equation}
    \text{threshold} = \arg\max_{\theta} (\text{TPR}(\theta) - \text{FPR}(\theta))
\end{equation}

This principled threshold selection optimizes discrimination between ID samples (with reliable predictions) and OOD samples (with suspect predictions).

\subsection{OOD Detection via Reconstruction and Quantization Error Metrics}
Autoencoder models with discrete latent spaces inherently provide multiple signals that can be leveraged for out-of-distribution detection. We explore two complementary approaches that exploit different aspects of the model's behavior when confronted with unfamiliar data distributions.
The reconstruction capability of an autoencoder model represents a natural mechanism for OOD detection. While the decoder component is not directly utilized in our classification pipeline, it provides valuable diagnostic information regarding input normality. For an input sequence \(x\), we define the reconstruction error OOD score as:

\begin{equation}
    \mathcal{S}_{\text{recon}}(x) = \frac{1}{N} \sum_{i=1}^{N} ||x_i - \hat{x}_i||_2^2
\end{equation}

where \(\hat{x}\) represents the reconstructed sequence from the decoder and (N) is the sequence length. This metric effectively captures the model's inability to faithfully reproduce unfamiliar patterns, as reconstruction quality deteriorates significantly when the encoder-decoder pathway encounters representations that deviate from the training distribution. Higher values of \(S_{recon}\) correlate with increasing likelihood of OOD samples.
Complementing the reconstruction-based approach, the discrete quantization mechanism inherent to VQ-VAE architectures provides a second signal for detecting distribution shifts. During encoding, continuous latent vectors \(z_e(x)\) are mapped to their nearest neighbors in a learned discrete codebook \(e_k\) with \(k\) denoting the number of codebooks. 

The quantization error measures the discrepancy between these continuous encodings and their discrete counterparts:

\begin{equation}
    \mathcal{S}_{\text{quant}}(x) = \frac{1}{M} \sum_{j=1}^{M} ||z_e^j(x) - e_{k(j)}||_2^2
\end{equation}

where \(z_e^j(x)\) represents the (j)-th encoded vector before quantization, \(e_{k(j)}\) is the corresponding codebook vector selected through nearest-neighbor lookup, and (M) denotes the number of encoded vectors. This quantization error naturally increases when the model processes samples that generate embeddings in regions of the latent space poorly covered by the codebook vectors, thereby providing a computationally efficient indicator of distribution shift without requiring full sequence reconstruction.

\subsection{Autoregressive Loss for OOD Detection}
The model is trained on the autoregressive next-token prediction task and classification. The training regimen involved 30 epochs dedicated to the task with 2 intermediate classification epochs every 10th epoch. Afterwards, we fine-tuned the model on the classification task for five epochs.
The autoregressive component of our architecture enables the quantification of sequence probability distributions, providing a natural mechanism for OOD detection. For a discrete token \(sequence z=(z_1,z_2,\dots,z_T) \) derived from our VQ-VAE encoder, the autoregressive transformer models the conditional distribution \(p(z_t | z_1, \dots, z_{t-1})\) for each position \(t\). We leverage this property to define our OOD detection metric as:

\begin{equation}
    \mathcal{S}_{\text{AR}}(z) = -\frac{1}{T} \sum_{t=1}^{T} \log p(z_t | z_1, \dots, z_{t-1})
\end{equation}

This formulation represents the negative log-likelihood (NLL) of the entire sequence normalized by sequence length. The underlying principle is that sequences from the training distribution will exhibit lower NLL values, reflecting higher probability under the learned model, while OOD samples will generate significantly higher NLL values due to unexpected token transitions.

%% file: 06_experiments.tex
\section{Experiments}
\subsection{Out-of-Distribution Detection}
\begin{table*}[htb!]
\small
\caption{Model performance comparison. F1-score and Acc show classification performance; columns show \(\text{OOD-Score}_{\text{F1}}\) for different detection methods: \(\mathcal{S}_{\text{AR}}\), \(\mathcal{S}_{\text{recon}}\), \(\mathcal{S}_{\text{quant}}\), MSP (Maximum Softmax Probability), ODIN, Vac (Vacuity), Maha (Mahalanobis), and CODiT. Mean ± std over 5 seeds.}
\label{tab:model-performance-full}
\resizebox{\textwidth}{!}{%
\begin{tabular}{l|cc|ccc|ccccc}
\toprule
Model & F1-score & Acc & \(\mathcal{S}_{\text{AR}}\) & \(\mathcal{S}_{\text{recon}}\) & \(\mathcal{S}_{\text{quant}}\) & MSP & ODIN & Vac & Maha & CODiT\\
\midrule
CODiT & 0.44 ± 0.15 & \underline{0.58 ± 0.07} & - & - & - & - & - & - & - & 0.22 ± 0.06 \\
MLP & 0.52 ± 0.04 & 0.54 ± 0.03 & - & - & - & 0.28 ± 0.05 & 0.18 ± 0.03& 0.20 ± 0.04 & -0.00 ± 0.06 & - \\
MLP EDL & \underline{0.56 ± 0.11} & 0.57 ± 0.10 & - & - & - & 0.29 ± 0.07& 0.18 ± 0.04 & 0.23 ± 0.07& -0.01 ± 0.08 & -\\
VQ-VAE MLP & 0.50 ± 0.02 & 0.51 ± 0.02 & - & 0.20 ± 0.07 & 0.19 ± 0.04 & 0.14 ± 0.03 & 0.17 ± 0.02 & 0.17 ± 0.02 & -0.03 ± 0.01 & - \\
VQ-VAE Tr & \textbf{0.65 ± 0.03}& \textbf{0.65 ± 0.03}& \textbf{0.35 ± 0.08}& 0.28 ± 0.07& 0.20 ± 0.08& 0.30 ± 0.03& 0.25 ± 0.03& \underline{0.34 ± 0.02}& 0.05 ± 0.14 & - \\
\bottomrule
\end{tabular}
}
\end{table*}

We evaluated the proposed OOD detection approach against several established baselines across both welding quality prediction and OOD detection capabilities. The baseline models included: (i) CODiT, a temporal-based OOD detection framework leveraging conformity scores; (ii) a standard multi-layer perceptron (MLP); (iii) MLP with evidential deep learning (EDL); and (iv) a VQ-VAE-based MLP classifier. For OOD detection methods, we compared Maximum Softmax Probability (MSP), ODIN, Vacuity-based detection, and Mahalanobis distance approaches. Our primary model, the \ac{VQ-VAE-Tr}, was implemented following the architecture and training routine described in~\cite{hahnCIKM}.

We conducted a comprehensive hyperparameter search to optimize model configurations, evaluating multiple subsequence lengths spanning one to ten welding cycles as input sequences. For VQ-VAE-based architectures, a unified optimization strategy was employed where the best-performing VQ-VAE configuration was subsequently used across all downstream tasks—including both classification and OOD detection. Each experimental configuration was trained using five different random seeds to ensure statistical robustness. The reported mean and standard deviation values in Table \ref{tab:model-performance-full} reflect the performance across these five independent runs. The best hyperparameters and the full code are available on GitHub\footnote{\url{https://github.com/tmdt-buw/ood-detection-welding}}.

\subsection{OOD Results}
The results presented in Table \ref{tab:model-performance-full} demonstrate that the \ac{VQ-VAE-Tr} architecture achieves superior performance on the primary welding quality prediction task, achieving an F1-score that is 16\% higher than the second-best model. This represents the best classification performance of all the models evaluated.

The critical finding of our evaluation concerns the OOD detection capabilities across different approaches. The \ac{VQ-VAE-Tr} using autoregressive loss as the detection signal achieves the highest OOD detection F1-score of 0.35 ± 0.08, followed closely by the Vacuity-based approach at 0.34 ± 0.02. Notably, traditional reconstruction and encoder-based OOD detection methods within the VQ-VAE framework yielded substantially lower performance (\(\mathcal{S}_{\text{recon}}\)=0.28 ± 0.07 and \(\mathcal{S}_{\text{quant}}\)=0.20 ± 0.08, respectively). This phenomenon can be attributed to the VQ-VAE's generalization capabilities across distribution shifts, which manifest in robust reconstruction performance but fail to transfer to downstream classification tasks. While the encoder-decoder architecture demonstrates resilience to distributional variations, the classification head remains susceptible to these shifts, highlighting the differential impact of OOD data on distinct model components.

The superior performance of autoregressive loss-based OOD detection validates our hypothesis that predictive next-token uncertainty provides a more sensitive indicator of distribution shifts compared to reconstruction quality metrics. This advantage is particularly pronounced when compared to classical OOD detection approaches such as MSP (0.30 ± 0.03) and ODIN (0.25 ± 0.03), which rely on output confidence calibration rather than learned sequence dynamics.

Moreover, the Mahalanobis distance approach consistently underperformed across all model architectures, suggesting that simple distance-based metrics in feature space are insufficient for capturing the complex distributional shifts present in welding time series data.

\subsection{Continual Learning in a Welding Process}
To evaluate the practical applicability of our OOD detection approach, we tested a realistic deployment scenario that mimics the evolving nature of industrial welding processes. We implemented a deployment simulation that reflects a typical production cycle where a quality prediction model initially trained on one welding configuration encounters gradually shifting process distributions. The scenario progressively introduces new experiences—consisting of additional welding runs with novel welding parameter combinations from the different joint types.

Our experimental framework evaluates three distinct deployment strategies:
First, we establish a baseline by deploying the static model without any adaptation capability. This approach represents the common industrial practice where models remain fixed after initial training, providing insight into performance degradation under natural distribution shifts.

Second, we implement continuous adaptation using experience replay~\cite{rolnick2019experience}, a continual learning strategy particularly well-suited for manufacturing where historical welding data remains accessible and provides stable adaptation for binary quality classification. In this configuration, the model continuously updates its parameters whenever new labeled data becomes available, regardless of whether the data represents a significant distribution shift.

Third, we introduce our OOD-triggered adaptation approach, where the autoregressive loss-based OOD detection mechanism serves as a gatekeeper for model updates. The system continuously monitors incoming data streams, triggering experience replay adaptation only when the OOD detector identifies samples exceeding the established threshold. This selective updating strategy aims to balance adaptation needs with computational efficiency and labeling costs.

\begin{figure}
    \centering
    \includegraphics[width=0.8\linewidth]{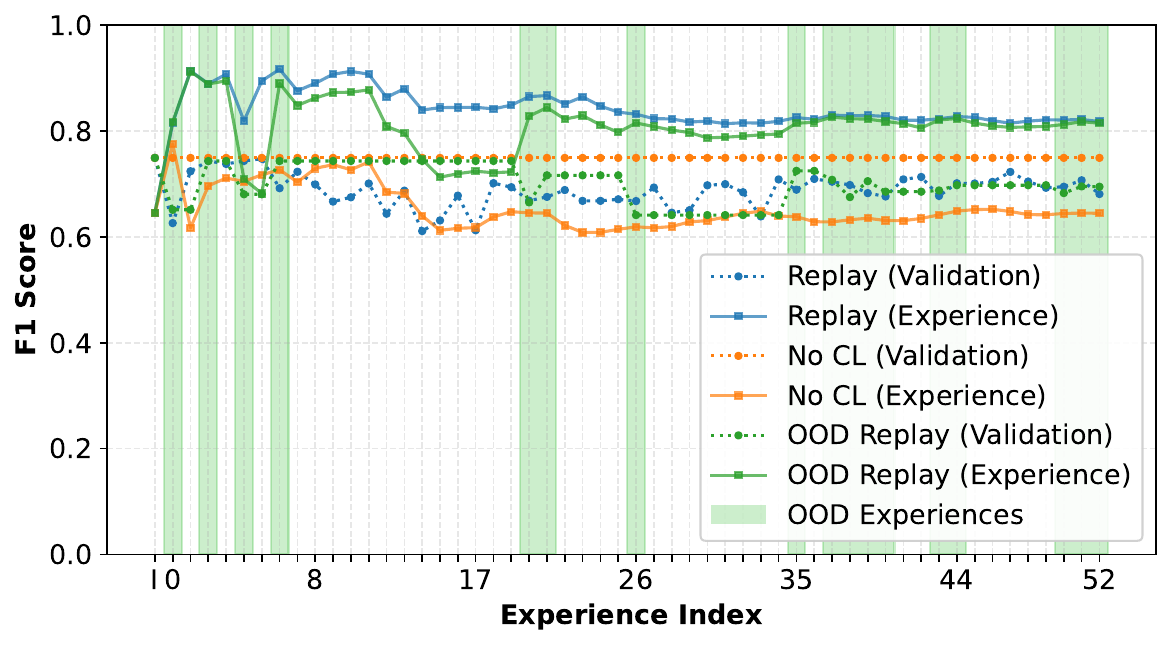}
    \caption{Continual learning performance over 53 sequential experiences comparing three deployment strategies: static baseline (No CL), continuous adaptation (Replay), and OOD-triggered selective adaptation (OOD Replay). Green regions indicate OOD detection triggers and the \textit{I} standing for the initial performance.}
    \label{fig:ood_continual_learning_results}
\end{figure}

The results presented in Figure~\ref{fig:ood_continual_learning_results} demonstrate the effectiveness of our OOD-triggered continual learning approach. The experimental evaluation spans 53 sequential experiences, with green shaded regions indicating instances where the OOD detection mechanism triggers model adaptation.

Our selective adaptation strategy achieves a 67.9\% reduction in labeling requirements, requiring updates at only 17 out of 53 experiences while maintaining performance comparable to continuous adaptation. The OOD-triggered approach (OOD Replay) exhibits two notable characteristics: (i) transient performance degradation during initial distribution shifts during experiences 10-15, reflecting the model's adaptation to novel welding configurations, and (ii) after experience 20 subsequent convergence to performance levels approximately equivalent to full continual learning (Replay). The static baseline (No CL) demonstrates progressive performance deterioration of the experience F1-score, underscoring the necessity of adaptation mechanisms in evolving manufacturing environments.

%% file: 07_conclusion_outlook.tex
\section{Conclusion and Outlook}
In this work, we addressed a fundamental challenge in industrial machine learning: maintaining robust predictive performance while minimizing costly labeling requirements as manufacturing processes evolve. We present a novel approach to out-of-distribution detection in dynamic welding processes, demonstrating that autoregressive loss from a \ac{VQ-VAE-Tr} architecture provides a reliable mechanism for identifying distribution shifts in welding quality prediction tasks. By integrating OOD detection with selective continual learning.

This work makes several key contributions to the field of adaptive quality monitoring in manufacturing. First, we establish that autoregressive loss serves as a superior OOD indicator compared to conventional reconstruction or quantization error-based approaches, achieving an \(OOD_{F1-score}\) of 0.35 ± 0.08 for OOD detection — outperforming traditional methods such as MSP and ODIN. Second, we introduce a quantitative OOD score metric that simultaneously evaluates a model's in-distribution performance and its ability to distinguish OOD samples, providing a unified framework for assessing model reliability. Third, we demonstrate that OOD-triggered continual learning maintains quality prediction accuracy comparable to continuous adaptation while substantially reducing the labeling requirements by 67.9\%, addressing a critical cost consideration in industrial deployments. This efficiency gain is particularly important in manufacturing contexts where obtaining quality labels requires destructive testing procedures.

By enabling autonomous detection and adaptation to process changes, this framework paves the way for self-maintaining quality systems that remain robust across welding configurations, parameter drifts, and unexpected distribution shifts. Future research directions include extending the approach to multi-modal sensor data, investigating more sophisticated continual learning strategies that leverage the discrete token representations, and exploring the application of this framework to other manufacturing processes beyond welding. While demonstrated on the \ac{VQ-VAE-Tr}, the core principle—leveraging autoregressive loss as an OOD indicator to trigger selective adaptation—could potentially enhance other autoregressive models and manufacturing domains where temporal patterns encode critical process dynamics.

%% file: 08_gen_ai_statement.tex
\section*{GenAI Usage Disclosure}

Generative AI tools were used to support code development and assist in drafting parts of the text. All content was critically reviewed, edited, and remains the sole responsibility of the authors.

%% file: ref.bib
@inproceedings{zhouPredictingQualityAutomated2020,
  title = {Predicting {{Quality}} of {{Automated Welding}} with {{Machine Learning}} and {{Semantics}}: {{A Bosch Case Study}}},
  shorttitle = {Predicting {{Quality}} of {{Automated Welding}} with {{Machine Learning}} and {{Semantics}}},
  booktitle = {Proceedings of the 29th {{ACM International Conference}} on {{Information}} \& {{Knowledge Management}}},
  author = {Zhou, Baifan and Svetashova, Yulia and Byeon, Seongsu and Pychynski, Tim and Mikut, Ralf and Kharlamov, Evgeny},
  year = {2020},
  month = oct,
  pages = {2933--2940},
  publisher = {{ACM}},
  address = {{Virtual Event Ireland}},
  doi = {10/gst793},
  urldate = {2023-10-11},
  isbn = {978-1-4503-6859-9},
  langid = {english},
}

@article{Bestard.2018,
 author = {Bestard, Guillermo Alvarez and {Absi Alfaro}, Sadek Crisstomo},
 doi = {10/jqw6},
 issn = {1678-5878},
 journal = {Journal of the Brazilian Society of Mechanical Sciences and Engineering},
 number = {9},
 title = {Measurement and estimation of the weld bead geometry in arc welding processes: the last 50 years of development},
 volume = {40},
 year = {2018}
}

@article{Fan.2022,
 author = {Fan, Shilong and Yang, Fei and Zhu, Xiaonan and Diao, Zhaowei and Chen, Lin and Rong, Mingzhe},
 doi = {10/jqw9},
 issn = {1009-0630},
 journal = {Plasma Science {\&} Technology},
 number = {4},
 title = {Numerical analysis on the effect of process parameters on deposition geometry in wire arc additive manufacturing},
 volume = {24},
 year = {2022}
}

@article{Kesse.2020,
 author = {Kesse, Martin A. and Buah, Eric and Handroos, Heikki and Ayetor, Godwin K.},
 doi = {10/grhvxj},
 journal = {Metals},
 number = {4},
 title = {Development of an Artificial Intelligence Powered TIG Welding Algorithm for the Prediction of Bead Geometry for TIG Welding Processes using Hybrid Deep Learning},
 volume = {10},
 year = {2020}
}

@book{Weman.2003,
 address = {Cambridge Eng. and Boca Raton FL},
 author = {Weman, Klas},
 isbn = {1855736896},
 publisher = {{Woodhead Pub.} and {Published in North America by CRC Press}},
 title = {Welding processes handbook},
 year = {2003}
}

@article{Zhang.2021,
 author = {Zhang, Y. M. and Wang, Q. Y. and Liu, Y. K.},
 doi = {10/gn96d2},
 issn = {0043-2296},
 journal = {Welding Journal},
 number = {2},
 pages = {63S--83S},
 title = {Adaptive Intelligent Welding Manufacturing},
 volume = {100},
 year = {2021}
}

@article{hahn2023towards,
  title={Towards a Deep Learning-based Online Quality Prediction System for Welding Processes},
  author={Hahn, Yannik and Maack, Robert and Buchholz, Guido and Purrio, Marion and Angerhausen, Matthias and Tercan, Hasan and Meisen, Tobias},
  journal={Procedia CIRP},
  volume={120},
  pages={1047--1052},
  year={2023},
  publisher={Elsevier}
}

@misc{Pham.2022,
 author = {Pham, Quang and Liu, Chenghao and Sahoo, Doyen and Hoi, Steven C. H.},
 date = {2022},
 title = {Learning Fast and Slow for Online Time Series Forecasting},
 publisher = {arXiv},
 doi = {10.48550/arXiv.2202.11672}
}

@inproceedings{MustafaAbdool.2023,
 author = {{Mustafa Abdool} and {Andrew Nam} and {James McClelland}},
 title = {Continual Learning and Out of Distribution Generalization in a Systematic Reasoning Task},
 url = {https://openreview.net/forum?id=NdSGKZvX3z},
 booktitle = {The 3rd Workshop on Mathematical Reasoning and AI at NeurIPS'23},
 year = {2023}
}

@misc{Lu.2022,
 author = {Lu, Wang and Wang, Jindong and Sun, Xinwei and Chen, Yiqiang and Xie, Xing},
 date = {2022},
 title = {Out-of-Distribution Representation Learning for Time Series Classification},
 publisher = {arXiv},
 doi = {10.48550/arXiv.2209.07027}
}

@misc{Kim.2022,
 author = {Kim, Gyuhak and Ke, Zixuan and Liu, Bing},
 date = {2022},
 title = {A Multi-Head Model for Continual Learning via Out-of-Distribution Replay},
 publisher = {arXiv},
 doi = {10.48550/arXiv.2208.09734}
}

@misc{Kaur.2022,
 author = {Kaur, Ramneet and Sridhar, Kaustubh and Park, Sangdon and Jha, Susmit and Roy, Anirban and Sokolsky, Oleg and Lee, Insup},
 date = {2022},
 title = {CODiT: Conformal Out-of-Distribution Detection in Time-Series Data},
 publisher = {arXiv},
 doi = {10.48550/arXiv.2207.11769}
}

@article{Feng.2021,
 author = {Feng, Yeli and Ng, Daniel Jun Xian and Easwaran, Arvind},
 year = {2021},
 title = {Improving Variational Autoencoder based Out-of-Distribution Detection for Embedded Real-time Applications},
 pages = {1--26},
 volume = {20},
 number = {5s},
 issn = {1539-9087},
 journal = {ACM Transactions on Embedded Computing Systems},
 doi = {10.1145/3477026}
}

@misc{Du.2021,
 author = {Du, Yuntao and Wang, Jindong and Feng, Wenjie and Pan, Sinno and Qin, Tao and Xu, Renjun and Wang, Chongjun},
 date = {2021},
 title = {AdaRNN: Adaptive Learning and Forecasting of Time Series},
 publisher = {arXiv},
 doi = {10.48550/arXiv.2108.04443}
}

@misc{Doorenbos.2024,
 author = {Doorenbos, Lars and Sznitman, Raphael and M{\'a}rquez-Neila, Pablo},
 date = {2024},
 title = {Continual Unsupervised Out-of-Distribution Detection},
 publisher = {arXiv},
 doi = {10.48550/arXiv.2406.02327}
}

@misc{Aguilar.2023,
 author = {Aguilar, Eduardo and Raducanu, Bogdan and Radeva, Petia and {van de Weijer}, Joost},
 date = {2023},
 title = {Continual Evidential Deep Learning for Out-of-Distribution Detection},
 publisher = {arXiv},
 doi = {10.48550/arXiv.2309.02995}
}

@misc{Yang.2022,
 author = {Yang, Jingkang and Wang, Pengyun and Zou, Dejian and Zhou, Zitang and Ding, Kunyuan and Peng, Wenxuan and Wang, Haoqi and Chen, Guangyao and Li, Bo and Sun, Yiyou and Du, Xuefeng and Zhou, Kaiyang and Zhang, Wayne and Hendrycks, Dan and Li, Yixuan and Liu, Ziwei},
 date = {2022},
 title = {OpenOOD: Benchmarking Generalized Out-of-Distribution Detection},
 publisher = {arXiv},
 doi = {10.48550/arXiv.2210.07242}
}

@misc{he2022outofdistributiondetectionunsupervisedcontinual,
      title={Out-Of-Distribution Detection In Unsupervised Continual Learning}, 
      author={Jiangpeng He and Fengqing Zhu},
      year={2022},
      eprint={2204.05462},
      archivePrefix={arXiv},
      primaryClass={cs.CV},
      url={https://arxiv.org/abs/2204.05462}, 
}

@misc{hendrycks2018baselinedetectingmisclassifiedoutofdistribution,
      title={A Baseline for Detecting Misclassified and Out-of-Distribution Examples in Neural Networks}, 
      author={Dan Hendrycks and Kevin Gimpel},
      year={2018},
      eprint={1610.02136},
      archivePrefix={arXiv},
      primaryClass={cs.NE},
      url={https://arxiv.org/abs/1610.02136}, 
}

@inbook{928a56b7d6f1473e930f282a0c4b534e,
title = "A tutorial on energy-based learning",
author = "Yann Lecun and Sumit Chopra and Raia Hadsell and Ranzato, {Marc Aurelio} and Huang, {Fu Jie}",
year = "2006",
language = "English (US)",
editor = "G. Bakir and T. Hofman and B. Scholkopt and A. Smola and B. Taskar",
booktitle = "Predicting structured data",
publisher = "MIT Press",
}

@misc{liang2020enhancingreliabilityoutofdistributionimage,
      title={Enhancing The Reliability of Out-of-distribution Image Detection in Neural Networks}, 
      author={Shiyu Liang and Yixuan Li and R. Srikant},
      year={2020},
      eprint={1706.02690},
      archivePrefix={arXiv},
      primaryClass={cs.LG},
      url={https://arxiv.org/abs/1706.02690}, 
}

@inproceedings{10.1145/3580305.3599880,
author = {Tercan, Hasan and Meisen, Tobias},
title = {Online Quality Prediction in Windshield Manufacturing using Data-Efficient Machine Learning},
year = {2023},
isbn = {9798400701030},
publisher = {Association for Computing Machinery},
address = {New York, NY, USA},
url = {https://doi.org/10.1145/3580305.3599880},
doi = {10.1145/3580305.3599880},
booktitle = {Proceedings of the 29th ACM SIGKDD Conference on Knowledge Discovery and Data Mining},
pages = {4914–4923},
numpages = {10},
location = {Long Beach, CA, USA},
series = {KDD '23}
}

@misc{ramakrishna2021efficientoutofdistributiondetectionusing,
      title={Efficient Out-of-Distribution Detection Using Latent Space of $\beta$-VAE for Cyber-Physical Systems}, 
      author={Shreyas Ramakrishna and Zahra Rahiminasab and Gabor Karsai and Arvind Easwaran and Abhishek Dubey},
      year={2021},
      eprint={2108.11800},
      archivePrefix={arXiv},
      primaryClass={cs.LG},
      url={https://arxiv.org/abs/2108.11800}, 
}

@INPROCEEDINGS{9797620,
  author={Yang, Yahan and Kaur, Ramneet and Dutta, Souradeep and Lee, Insup},
  booktitle={2022 ACM/IEEE 13th International Conference on Cyber-Physical Systems (ICCPS)}, 
  title={Interpretable Detection of Distribution Shifts in Learning Enabled Cyber-Physical Systems}, 
  year={2022},
  volume={},
  number={},
  pages={225-235},
  doi={10.1109/ICCPS54341.2022.00027}}

@article{article,
author = {Tercan, Hasan and Deibert, Philipp and Meisen, Tobias},
year = {2022},
month = {01},
pages = {},
title = {Continual learning of neural networks for quality prediction in production using memory aware synapses and weight transfer},
volume = {33},
journal = {Journal of Intelligent Manufacturing},
doi = {10.1007/s10845-021-01793-0}
}

@misc{vandeven2019scenarioscontinuallearning,
      title={Three scenarios for continual learning}, 
      author={Gido M. van de Ven and Andreas S. Tolias},
      year={2019},
      eprint={1904.07734},
      archivePrefix={arXiv},
      primaryClass={cs.LG},
      url={https://arxiv.org/abs/1904.07734}, 
}

@inproceedings{hahnCIKM,
author = {Hahn, Yannik and Maack, Robert and Tercan, Hasan and Meisen, Tobias and Purrio, Marion and Buchholz, Guido and Angerhausen, Matthias},
title = {Quality Prediction in Arc Welding: Leveraging Transformer Models and Discrete Representations from Vector Quantised-VAE},
year = {2024},
isbn = {9798400704369},
publisher = {Association for Computing Machinery},
address = {New York, NY, USA},
url = {https://doi.org/10.1145/3627673.3680031},
doi = {10.1145/3627673.3680031},
booktitle = {Proceedings of the 33rd ACM International Conference on Information and Knowledge Management},
pages = {4522–4529},
numpages = {8},
location = {Boise, ID, USA},
series = {CIKM '24}
}

@article{youden1950index,
  title={Index for rating diagnostic tests},
  author={Youden, William J},
  journal={Cancer},
  volume={3},
  number={1},
  pages={32--35},
  year={1950},
  publisher={Wiley Online Library}
}

@article{rolnick2019experience,
  title={Experience replay for continual learning},
  author={Rolnick, David and Ahuja, Arun and Schwarz, Jonathan and Lillicrap, Timothy and Wayne, Gregory},
  journal={Advances in neural information processing systems},
  volume={32},
  year={2019}
}

@article{franc2023reject,
  title={Reject option models comprising out-of-distribution detection},
  author={Franc, Vojtech and Prusa, Daniel and Paplham, Jakub},
  journal={arXiv preprint arXiv:2307.05199},
  year={2023}
}

@article{tajwar2021no,
  title={No true state-of-the-art? ood detection methods are inconsistent across datasets},
  author={Tajwar, Fahim and Kumar, Ananya and Xie, Sang Michael and Liang, Percy},
  journal={arXiv preprint arXiv:2109.05554},
  year={2021}
}

@article{fort2021exploring,
  title={Exploring the limits of out-of-distribution detection},
  author={Fort, Stanislav and Ren, Jie and Lakshminarayanan, Balaji},
  journal={Advances in neural information processing systems},
  volume={34},
  pages={7068--7081},
  year={2021}
}

@inproceedings{humblot2023beyond,
  title={Beyond AUROC \& co. for evaluating out-of-distribution detection performance},
  author={Humblot-Renaux, Galadrielle and Escalera, Sergio and Moeslund, Thomas B},
  booktitle={Proceedings of the IEEE/CVF Conference on Computer Vision and Pattern Recognition},
  pages={3881--3890},
  year={2023}
}

@article{vapsi2025hypercone,
  title={Hypercone Assisted Contour Generation for Out-of-Distribution Detection},
  author={Vapsi, Annita and Mu{\~n}oz, Andr{\'e}s and Thomas, Nancy and Ramani, Keshav and Borrajo, Daniel},
  journal={arXiv preprint arXiv:2501.10209},
  year={2025}
}

@article{maschler2021regularization,
  title={Regularization-based continual learning for anomaly detection in discrete manufacturing},
  author={Maschler, Benjamin and Pham, Thi Thu Huong and Weyrich, Michael},
  journal={Procedia CIRP},
  volume={104},
  pages={452--457},
  year={2021},
  publisher={Elsevier}
}

@article{tercan2022continual,
  title={Continual learning of neural networks for quality prediction in production using memory aware synapses and weight transfer},
  author={Tercan, Hasan and Deibert, Philipp and Meisen, Tobias},
  journal={Journal of Intelligent Manufacturing},
  volume={33},
  number={1},
  pages={283--292},
  year={2022},
  publisher={Springer}
}

@inproceedings{10.1145/3340531.3412737,
author = {Zhou, Baifan and Svetashova, Yulia and Byeon, Seongsu and Pychynski, Tim and Mikut, Ralf and Kharlamov, Evgeny},
title = {Predicting Quality of Automated Welding with Machine Learning and Semantics: A Bosch Case Study},
year = {2020},
isbn = {9781450368599},
publisher = {Association for Computing Machinery},
address = {New York, NY, USA},
url = {https://doi.org/10.1145/3340531.3412737},
doi = {10.1145/3340531.3412737},
booktitle = {Proceedings of the 29th ACM International Conference on Information \& Knowledge Management},
pages = {2933–2940},
numpages = {8},
location = {Virtual Event, Ireland},
series = {CIKM '20}
}

@inproceedings{10.1145/3511808.3557512,
author = {Zheng, Zhuoxun and Zhou, Baifan and Zhou, Dongzhuoran and Soylu, Ahmet and Kharlamov, Evgeny},
title = {Executable Knowledge Graph for Transparent Machine Learning in Welding Monitoring at Bosch},
year = {2022},
isbn = {9781450392365},
publisher = {Association for Computing Machinery},
address = {New York, NY, USA},
url = {https://doi.org/10.1145/3511808.3557512},
doi = {10.1145/3511808.3557512},
booktitle = {Proceedings of the 31st ACM International Conference on Information \& Knowledge Management},
pages = {5102–5103},
numpages = {2},
location = {Atlanta, GA, USA},
series = {CIKM '22}
}

@article{GRAHAM2023102967,
title = {Latent Transformer Models for out-of-distribution detection},
journal = {Medical Image Analysis},
volume = {90},
pages = {102967},
year = {2023},
issn = {1361-8415},
doi = {https://doi.org/10.1016/j.media.2023.102967},
url = {https://www.sciencedirect.com/science/article/pii/S136184152300227X},
author = {Mark S. Graham and Petru-Daniel Tudosiu and Paul Wright and Walter Hugo Lopez Pinaya and Petteri Teikari and Ashay Patel and Jean-Marie U-King-Im and Yee H. Mah and James T. Teo and Hans Rolf Jäger and David Werring and Geraint Rees and Parashkev Nachev and Sebastien Ourselin and M. Jorge Cardoso}
}

@inproceedings{NEURIPS2022_3066f60a,
 author = {Li, Yewen and Wang, Chaojie and Xia, Xiaobo and Liu, Tongliang and miao, xin and An, Bo},
 booktitle = {Advances in Neural Information Processing Systems},
 editor = {S. Koyejo and S. Mohamed and A. Agarwal and D. Belgrave and K. Cho and A. Oh},
 pages = {7383--7396},
 publisher = {Curran Associates, Inc.},
 title = {Out-of-Distribution Detection with An Adaptive Likelihood Ratio on Informative Hierarchical VAE},
 url = {https://proceedings.neurips.cc/paper_files/paper/2022/file/3066f60a91d652f4dc690637ac3a2f8c-Paper-Conference.pdf},
 volume = {35},
 year = {2022}
}

@inproceedings{NEURIPS2020_eddea82a,
 author = {Xiao, Zhisheng and Yan, Qing and Amit, Yali},
 booktitle = {Advances in Neural Information Processing Systems},
 editor = {H. Larochelle and M. Ranzato and R. Hadsell and M.F. Balcan and H. Lin},
 pages = {20685--20696},
 publisher = {Curran Associates, Inc.},
 title = {Likelihood Regret: An Out-of-Distribution Detection Score For Variational Auto-encoder},
 url = {https://proceedings.neurips.cc/paper_files/paper/2020/file/eddea82ad2755b24c4e168c5fc2ebd40-Paper.pdf},
 volume = {33},
 year = {2020}
}

@inproceedings{osada2023out,
  title={Out-of-distribution detection with reconstruction error and typicality-based penalty},
  author={Osada, Genki and Takahashi, Tsubasa and Ahsan, Budrul and Nishide, Takashi},
  booktitle={Proceedings of the IEEE/CVF Winter Conference on Applications of Computer Vision},
  pages={5551--5563},
  year={2023}
}

@article{nanaumi2024low,
  title={Low-Quality Image Detection by Hierarchical VAE},
  author={Nanaumi, Tomoyasu and Kawamoto, Kazuhiko and Kera, Hiroshi},
  journal={arXiv preprint arXiv:2408.10885},
  year={2024}
}

@dataset{hahn_2025_15497262,
  author       = {Hahn, Yannik and
                  Königsfeld, Antonin and
                  Tercan, Hasan and
                  Buchholz, Guido and
                  Purrio, Marion and
                  Angerhausen, Matthias and
                  Meyes, Richard and
                  Meisen, Tobias},
  title        = {Metal Arc Welding},
  month        = may,
  year         = 2025,
  publisher    = {Zenodo},
  doi          = {10.5281/zenodo.15497262},
  url          = {https://doi.org/10.5281/zenodo.15497262},
}
